\documentclass[runningheads]{llncs}
\usepackage{graphicx}
\usepackage{comment}
\usepackage{amsmath,amssymb} % define this before the line numbering.
\usepackage{color}
\usepackage{times}
\usepackage[ruled,vlined]{algorithm2e}
\usepackage{amsfonts}
\usepackage{booktabs}  
\usepackage{siunitx}
\usepackage{floatrow}
\usepackage{subfig}
\usepackage{caption}

\begin{document}
% \linenumbers
\pagestyle{headings}
\mainmatter

\title{Self-Supervised Contextual Bandits in Computer Vision} % Replace with your title

\begin{comment}
\titlerunning{Self-Supervised Contextual Bandits in Computer Vision} 
\authorrunning{Deshmukh, Kumar, Boyles, Charles, Manavoglu, Dogan} 
\author{Aniket Anand Deshmukh, Abhimanu Kumar, Levi Boyles, Denis Charles, Eren Manavoglu, Urun Dogan}
\institute{Microsoft Ads, Microsoft AI \& Research}
\end{comment}
%******************

%\begin{comment}
\titlerunning{Self-Supervised Contextual Bandits in Computer Vision}
\author{Aniket Anand Deshmukh \and
Abhimanu Kumar \and
Levi Boyles \and
Denis Charles\and
Eren Manavoglu \and
Urun Dogan}
\authorrunning{Deshmukh et al.}
\institute{Microsoft Ads, Microsoft AI \& Research \\
\email{aniketde@umich.edu}}
%\end{comment}
%******************
\maketitle

\begin{abstract}
Contextual bandits are a common problem faced by machine learning practitioners in domains as diverse as hypothesis testing to product recommendations. There have been a lot of approaches in exploiting rich data representations for contextual bandit problems with varying degree of success. Self-supervised learning is a promising approach to find rich data representations without explicit labels. In a typical self-supervised learning scheme, the primary task is defined by the problem objective (e.g. clustering, classification, embedding generation etc.) and the secondary task is defined by the self-supervision objective (e.g. rotation prediction, words in neighborhood, colorization, etc.). In the usual self-supervision, we learn implicit labels from the training data for a secondary task. However, in the contextual bandit setting, we don't have the advantage of getting implicit labels due to lack of data in the initial phase of learning. We provide a novel approach to tackle this issue by combining a contextual bandit objective with a self supervision objective. By augmenting contextual bandit learning with self-supervision we get a better cumulative reward. Our results on eight popular computer vision datasets show substantial gains in cumulative reward. We provide cases where the proposed scheme doesn't perform optimally and give alternative methods for better learning in these cases.
\keywords{Contextual bandits, Self Supervised Learning}
\end{abstract}

\section{Introduction}

Allocating limited resources between competing choices when the choice attributes are only partially available or not available is a classic multi-armed bandit problem. For example: choosing an ad to display out of many possible ads given a user visit to a website, or recommending a news article out of many possible articles to a visiting reader. For choosing an ad to display, we only have partial labels even when a user clicks on the displayed ad since we don't observe whether the user would have clicked or not on non-displayed ads. The system needs to decide between displaying the current best performing set of ads (exploitation) and trying to discover new ads that can perform as well or better (exploration). Multi-armed bandit settings occur naturally in sequential decision process like these where there is an explore and exploit trade off. Traditionally a multi-armed bandit problem is solved using  $\epsilon$-greedy~\cite{Tokic_10_1007}, UCB based~\cite{pmlr-v22-kaufmann12}, Thompson Sampling (or Bayesian methods)~\cite{agrawal2013thompson,riquelme2018deep}, bootstrap based approach~\cite{kveton2019garbage}, exp4-type approach~\cite{McMahan_35401} etc. All of these methods have a combination of explore and exploit phases in the algorithm with varying degree of exploration and exploitation. 

Compared to multi-armed bandits, the contextual bandit setting has features that define the context of each action or arm. Contextual bandits are one of the most common bandit settings encountered in real world. E.g. recommending an ad or a news article to a user has features such as demography, location, webpage metadata etc. An ad (or arm) to be chosen for serving in order to obtain a click (or reward) will be based on these features (or context).
The bandit reward to be optimized in this case will be a function of these features. This is the underlying principal of many of the contextual bandit solvers such as LinUCB~\cite{chu2011contextual}, KernelUCB~\cite{kveton2019garbage}, Thompson sampling~\cite{agrawal2013thompson} etc.  The contextual bandit setting also solves the cold start problem quite prevalent in recommendation engines~\cite{li2016collaborative}. In case of a new user without history, the bandit approach is to initially rely more on the exploration phase. This simple approach results in relatively good gains.

Representation learning in contextual bandits setting has been studied in recent times via the use of deep neural networks~\cite{riquelme2018deep}. A good feature representation learning can lead to good gains in reward optimization in bandit tasks. \cite{riquelme2018deep} shows that using neural network with Bayesian linear regression as the top layer helps contextual bandits setting learn good representations. This in turn leads to better bandit reward optimization. Self-supervision is another approach for learning good feature representation for machine learning tasks~\cite{dosovitskiy2015discriminative,gidaris2018unsupervised,noroozi2016unsupervised,zhang2016colorful,goyal2019scaling,pathak2017curiosity,burda2018large,pathak2019self}. Given the label scarcity issue faced by machine learning practitioners, self-supervision has been used as one of the primary methods to leverage unlabeled data for supervised machine learning problems~\cite{dosovitskiy2015discriminative,gidaris2018unsupervised,noroozi2016unsupervised,zhang2016colorful}. Self-supervised learning combines unlabeled data with implicit labeling~\cite{mikolov2013efficient} to find rich feature representations~\cite{saunshi2019theoretical}. The richer representations help in the overall supervised or unsupervised task at hand. Self-supervised approaches have led to remarkable gains in label scarce tasks---image recognition ~\cite{taha2018stream,Singh2018SelfSupervisedFL}, clustering~\cite{Caron_2018_ECCV}, classification~\cite{10.1145/1645953.1646072}, few-shot learning~\cite{Gidaris_2019_ICCV}, semi-supervised learning~\cite{rebuffi2019semisupervised}, learning to rank~\cite{pami/LiuWB19} etc. Self-supervised learning is a promising approach to solving the label scarcity issue especially if the input space has good representational structure. 

UCB (upper confidence bound) based contextual bandit solvers select arms that maximize an acquisition function which is based on an estimated UCB of the reward. LinUCB finds a linear relationship between the current expected reward of an arm and all its previous rewards. The feature vectors in current round are reformulated as a linear combination of feature vectors in previous rounds. These computed linear weights and rewards in the previous rounds are used to compute the reward in the current round~\cite{chu2011contextual}. Rather than use a linear model, nonlinear methods may be used that extend the LinUCB approach by finding a non-linear weight combination using a deep neural network---we call this Neural-UCB. 

In this work we combine Neural-UCB with self-supervised learning to utilize the rich representation extracted by self-supervision. The self supervision helps by finding better feature representations given that the data labels are scarce for bandits setting especially in the exploration phase. We demonstrate the effectiveness of this combination on image classification dataset in a contextual bandits setting. Image ads are a common ad entity. Image ads recommendation tasks subsume image classification task since an ad-recommendation engine has to know the content of the image in the ad before showing it to the user. We convert the image classification problem into a contextual multi-arm bandits problem by passing sequence of images to a bandits setting and rewarding it for guessing the correct label (max-reward action or pulling the correct arm of the bandit) of the image. Our primary contributions are: 
\begin{itemize}
    \item We propose a novel technique to combine self-supervision with contextual bandits objective that provides good gains in the cumulative reward optimization. This is the first ever technique proposed for leveraging representation learning (via self-supervision) in contextual bandits task, as far as the authors know.
    \item We analyze the proposed method over eight diverse real world datasets showing good empirical gains across multiple random runs of the proposed algorithm.
    \item We empirically show the substantial benefits of self-supervision and scenarios where it helps the most in contextual bandits setting. 
\end{itemize}

\section{Motivation}
The multi-armed bandit setting suffers from data and label scarcity (or limited feedback) issues especially in the initial phase of the algorithm. In this phase, exploration is more prominent than exploitation, as the model needs more labels and data for accurate predictions. Recently self-supervision techniques have shown remarkable accuracy improvements in supervised and unsupervised label scarce tasks~\cite{taha2018stream,Singh2018SelfSupervisedFL,Caron_2018_ECCV,10.1145/1645953.1646072,Gidaris_2019_ICCV,rebuffi2019semisupervised,pami/LiuWB19}. The gains in self-supervision primarily come from extracting a better feature representation of the data using implicit labels. It has been shown that good representation learning can help multi-armed bandits setting~\cite{riquelme2018deep}. Our aim is to help contextual bandits' data and label scarcity issue by leveraging self-supervision scheme in getting better feature representation of the data.

We drive the effectiveness self-supervision in contextual bandits setting via demonstrating its superior performance in sequential image classification. Contextual bandits are a common approach for solving ads recommendation as they alleviate the cold start issue that the recommendation task suffers from~\cite{li2016collaborative}. Image ads are a common ad type where image classification is an important sub-task of image ad recommendation. An image ad recommendation system needs to know the semantic content of an image ad. In this work, we formulate image classification in a contextual bandits setting. The contextual bandit is provided a sequence of images and predicting the correct class (equivalent to pulling the most rewarding arm) leads to a reward of 1 otherwise 0. In this problem setting, we demonstrate the effectiveness of augmenting contextual bandits solver with self-supervised scheme to improve cumulative rewards optimization. We demonstrate the effectiveness of rich feature representations extracted via self-supervision in solving contextual bandits tasks.

\section{Contextual Bandits and Self-Supervision}
\label{sec:bandits_ss}
\begin{algorithm}
		\DontPrintSemicolon
		\caption{Contextual Bandits Problem}
		\label{alg:CMAB}
		%\textbf{Input:} $ \alpha \in R_+, $ \;
		\For{$t = 1,...,T$}
		{
	 Observe context $x_{a,t} \in \mathbb{R}^d$ for all arms $a \in [N]$, where $ [N] = \{1,...N\}$  \;
	  Choose an arm $a_t \in [N]$  \; %based on $(x_{a_{tau},\tau} , a_{\tau} , r_{a_{\tau},\tau})$
		 Receive a reward $r_{a_t,t} \in  \mathbb{R}$  \;
		 Improve arm selection strategy based on new observation $(x_{a_t,t} , a_t , r_{a_t,t})$ \;
		}
		Goal: Maximize cumulative reward := $ \sum_{t=1}^T r_{a_t,t}$
\end{algorithm}
Contextual Bandits are a class of bandit algorithms wherein, at round $t$, each arm $a_t$ is associated with a context $x_{a_t,t}$.  In such problems, rewards are assumed dependent on these observed covariates, which allow for conditional modelling of the reward, that is, estimation of $p(r_{a_t,t} | x_{\cdot, t}, a_t)$.  As in multi-armed bandits, contextual bandits are used for situations where standard supervised learning methods may be asymptotically suboptimal by failing to explore optimal arms enough.  This exploration is typically related to (estimates of) the variance of the estimator $\hat{p}(r_{a_t,t} | x_{\cdot,t}, a_t)$.

Due to the conditional dependence on $x$, some assumptions about $p(r_{a_t,t} | x_{\cdot, t}, a_t)$ are required in order to expect a vanishing uncertainty over time, so that the bandit algorithm can enter an exploitation phase and achieve sublinear regret.  One such assumption is a linear assumption on the relationship between $r_{a_t,t}$ and $x_{a_t,t}$ as in LinUCB \cite{chu2011contextual}.  Bandit algorithms with nonlinear dependencies include GP-UCB \cite{srinivas2010gaussian} and KernelUCB \cite{valko2013finite}, which assume dependencies drawn from Gaussian Process, or lying in a Reproducing Kernel Hilbert Space, respectively.  These three methods utilize the Upper Confidence Bound (UCB) as the means of uncertainty quantification for the purposes of trading off between exploration and exploitation.  Other methods for handling uncertainty can be used in contextual bandits, for example Thompson Sampling \cite{agrawal2013thompson} and the Bootstrapping \cite{kveton2019garbage}. Algorithm \ref{alg:CMAB} illustrates our contextual bandit setting and the goal is to maximizing the cumulative reward.  

Contextual bandits have been used extensively in sequential decision making tasks. \cite{li2011unbiased} elaborates an evaluation scheme for contextual bandits based recommendation engines.~\cite{li2016collaborative} proposes a contextual bandits based dynamic clustering approach to collaborative filtering that reduces the effect of classic cold-start problem using explore and exploit scheme. It shows good gains on diverse set of recommendation task datasets.~\cite{tewari2017ads} provides modifications needed in contextual bandit settings to adapt it for mobile health tasks.~\cite{deshmukh2018simple} devises a scheme where the bandit solver, called Contextual-Gap algorithm, has a pure exploration phase (feedback received but no regrets incurred) and a pure exploitation phase (regret incurred but no feedback). The Contextual-Gap algorithm shows good gains in regret minimization for adaptive sensor selection for magnetic field estimation in interplanetary spacecraft. 

\begin{algorithm}
		\DontPrintSemicolon
		\caption{Standard Self-Supervised Learning Problem}
		\label{alg:SSS}
		%\textbf{Input:} $ \alpha \in R_+, $ \;
		Get the unsupervised data \;
		Build a self-supervised task \;
		Learn representation of original data using defined self-supervised task \;
		Use the learnt representation for downstream tasks
\end{algorithm}

We combine algorithm \ref{alg:CMAB} with self-supervision to help the data scarcity issues as mentioned earlier. Given correct model specification and enough labelled data, supervised learning methods supply decent performance on the task. 
The core idea of self-supervision is to create an auxiliary task for which one can easily generate labels and then train a model on this supervised auxiliary task. Traditionally, after the model is trained on auxiliary (self-supervised) task, we use trained model as a feature extractor and train a linear predictor for the final machine learning task as illustrated in algorithm \ref{alg:SSS}. Although summarizing all self-supervised methods is beyond the scope of this paper, we will briefly summarize some of them in order to present generic idea of self-supervision. For all these methods, we will assume that a seed data set $\mathcal{D}_s$ is given.Exemplar CNN \cite{dosovitskiy2015discriminative}: For a subset of $\mathcal{D}_s$ of size $r$, first generate patches of a predefined size and then apply $k$ different transformations to these patches. The self-supervised task is to classify transformed images in to $r$ classes. Rotation \cite{gidaris2018unsupervised}: Each image in $\mathcal{D}_s$ is rotated by $90, 180, 270$ and $360$ degrees i.e. multiples of $90$ degrees. The self-supervised task is to classify rotated images in to $4$ classes. Jigsaw Puzzle \cite{noroozi2016unsupervised}: An image is divided in to 9 regions by using $3\times 3$ grid and these regions are shuffled. The self-supervised task is to predict the correct location of each region given the shuffled regions.

Self-supervised learning methods have also been used in reinforcement learning (RL) settings~\cite{pathak2017curiosity,burda2018large,pathak2019self}. One of the state-of-the art methods in RL is using self-supervision to increase the exploration of unpredictable states given the current state \cite{pathak2017curiosity,burda2018large}. Another prominent method is to train a set of agents and explore the states with maximum disagreement \cite{pathak2019self}. Self-supervision in RL leverages the fact that representation learning can lead to predicting next state given the current state which is not the case in contextual bandits. Hence, self-supervised methods show impressive performance on supervised learning tasks and improve the state of the art in RL settings, it is not obvious whether they can be used in contextual bandit settings directly.
\begin{algorithm}
		\DontPrintSemicolon
		\caption{Lin-UCB}\label{alg:LinUCB}
		\KwIn{Input: $ \alpha, \lambda \in R_+ $}
		\textbf{Initialize:} $ D_a = \lambda I_d $, $ b_a = 0_d $ and $\hat{\theta}_a = D_a^{-1}b_a $ for each $a \in [N]$  \;
		\For{$t = 1,...,T$}
		{
		 Observe context features at time $t$: $x_{a,t} $ for each $a \in [N]$. \;
		\For{all $a$ at time $t$}
		{
	   $\hat{r}_{a,t} = \hat{\theta}_{a}^Tx_{a,t} $ \;
	   $ s_{a,t} = \sqrt{x_{a,t}^TD_a^{-1}x_{a,t}}$ \;
	   $p_{a,t} \leftarrow   \hat{r}_{a,t} + \alpha s_{a,t}$ \;
		}  \;
		Choose arm $ a_t = \arg \max p_{a,t}$, observe a real valued payoff $ r_{a_t,t}$ and update $ y_t$ .\;
		Update $D_{a_t} = D_{a_t}  + x_{a_t,t} x_{a_t,t}^T $\;
		Update $b_{a_t} = b_{a_t}  + r_{a_t,t}x_{a_t,t} $\;
		Update $\hat{\theta}_{a_t} = D_{a_t}^{-1}b_{a_t}$\;
		\KwOut{Output: $ a_t $} \;
		}
\end{algorithm}
\begin{algorithm}
		\DontPrintSemicolon
		\caption{Neural-UCB}\label{alg:NeuralUCB}
		\KwIn{Input: $ \alpha, \lambda \in R_+ $}
		\textbf{Initialize:} $ D_a = \lambda I_{d'} $, $ b_a = 0_{d'} $ and $\hat{\theta} = D_a^{-1}b_a $ for each $a \in [N]$  \;
		\textbf{Initialize:} Neural Network $f_1: \mathbb{R}^d \times \mathbb{R}^{d'}$  \;
		\For{$t = 1,...,T$}
		{
		 Observe context features at time $t$: $x_{a,t} $ for each $a \in [N]$. \;
		 Get representation of context features at time $t$ using neural network $f(x_{a,t})$ for each $a \in [N]$. \;
		\For{all $a$ at time $t$}
		{
	   $\hat{r}_{a,t} = \hat{\theta}_a^Tf_t(x_{a,t}) $ \;
	   $ s_{a,t} = \sqrt{f_t(x_{a,t})^TD_a^{-1}f_t(x_{a,t})}$ \;
	   $p_{a,t} \leftarrow   \hat{r}_{a,t} + \alpha s_{a,t}$ \;
		}  \;
		Choose arm $ a_t = \arg \max p_{a,t}$, observe a real valued payoff $ r_{a_t,t}$ and update $ y_t$ .\;
		Update history  $H_t = (x_{a_t,t} , a_t , r_{a_t,t})$ \;
		\If{ $t \mod \beta == 0$}
		{
		   Retrain Neural Network $f_{t}$ using history $H_t$ to get $f_{t+1}$    \;
		}
		\Else
		{
		$f_{t+1} = f_{t}$
		}
		Update $D_{a_t} = \sum_{\tau }  f_{t+1}(x_{a_\tau,\tau})  f_{t+1}(x_{a_\tau,\tau})^T $\;
		Update $b_{a_t} =\sum_{\tau }  r_{a_\tau,\tau} f_{t+1}(x_{a_\tau,\tau}) $\;
		Update $\hat{\theta}_{a_t} = D_{a_t}^{-1}b_{a_t}$\;
		\KwOut{Output: $ a_t $} \;
		}
\end{algorithm}

\section{Proposed Method}
We first describe the existing contextual bandit solvers, specifically, Lin-UCB and Neural-UCB. Later, we elaborate the proposed approach of combining self-supervision and contextual bandits, SS-Neural-UCB. We do not add self-supervision to Lin-UCB as Lin-UCB does not leverage feature representation learning as is the case with Neural-UCB. 
\paragraph{\textbf{Lin-UCB}: }
The Lin-UCB, algorithm \ref{alg:LinUCB}, assumes a linear model for the reward: $\hat{r}_a = \theta_a x_a$.  One benefit of this approach is that the linear least squares (LLS) estimates $\hat{\theta}_a$ may be constructed in an online fashion.  Specifically, recall that the LLS estimator for design matrix $X$ and response vector $R$ is:
\begin{equation}
    \hat{\theta} = (X^TX)^{-1} X^TR  
\end{equation}
%Define X and R
By performing an online update for $b = X^T R$ and rank-one updates of a scatter matrix $D = X^TX$, estimates for each arm $\hat{\theta}_a$ may be efficiently updated when observing new data.  Furthermore, the LLS perspective allows one to estimate the variance of the predicted reward at a particular context $x$ as:
\begin{equation}
    Var(\hat{r}) = Var(\hat{\theta}^T x) = x^T Var(\hat{\theta}) x \approx x^T D^{-1} x \sigma^{2}.
\end{equation}
Here $\sigma^2$ is the variance of the (assumed homoscedastic) noise, and $Var(\hat{\theta})\approx \sigma^2 D^{-1}$ can be derived from the fact that $\hat{\theta}$ is a maximum likelihood estimator (and thus asymptotically efficient) and its variance is inversely related to the Fisher information.

This estimated variance is then used to construct an upper confidence bound (UCB) for each arm, and the arm is chosen by maximizing the UCB -- see Algorithm \ref{alg:LinUCB}.
\paragraph{\textbf{Neural-UCB:} }
The Neural-UCB, algorithm \ref{alg:NeuralUCB}, operates in a similar to Lin-UCB, except that a nonlinear model $f$ is used to construct an intermediate representation for the context $x$. This representation of the context $x$ is then used in place of the real $x$ in Lin-UCB for estimation of $\hat{r}$ and its variance.  Note that the full history is retained for training $f$.  Introducing the nonlinear layer gives the bandit algorithm the additional flexibility required to model arbitrarily complex tasks while still doing trade off between exploration and exploitation. 

\subsection{SS-Neural-UCB (Proposed Method)}
The nonlinear mapping $f$ learnt by Neural-UCB is critical to the performance of the bandit algorithm.  In the early stages of training, these complex representations may not see enough data in order to produce good predictions.  This either leads to suboptimal performance, or requires careful tuning of learning rates and regularization decay.  One successful approach in computer vision for learning complex and useful representations from limited data is self-supervision (see Section \ref{sec:bandits_ss}) which we propose to leverage in training Neural-UCB bandit algorithms.
% in order to learn better representations from limited data.
Our proposed bandit algorithm (algorithm~\ref{alg:SSNeural}) combines the bandit loss with self-supervised loss:
\begin{equation}
\label{eq:loss}
    Loss = L_b + (\mu)^{\frac{t}{\beta}} L_{ss},
\end{equation}
where $L_b$ is the bandits loss (least squares), $L_ss$ is the self-supervised loss (auxiliary task loss), $\mu \in [0,1)$ is the weight on self-supervised loss and $t$ is the time step. We are decaying the weight given to self-supervision since the bandits setting needs more guidance in the beginning phases (dominated by exploration and data scarcity).  In case of $\mu = 1 $, the weight on self-supervised loss never goes to zero and in that case self-supervised loss may give sub-optimal performance on the original bandit task, as after sufficient rounds the network model has enough data to learn good representations. We show experimental results for $\mu = 1 $ in appendix.

\begin{algorithm}
		\DontPrintSemicolon
		\caption{SS-Neural-UCB}\label{alg:SSNeural}
		\KwIn{Input: $ \alpha, \beta \lambda \in R_+ $}
		\textbf{Initialize:} $ D_a = \lambda I_{d'} $, $ b_a = 0_{d'} $ and $\hat{\theta} = D_a^{-1}b_a $ for each $a \in [N]$  \;
		\textbf{Initialize:} Neural Network $f_1: \mathbb{R}^d \times \mathbb{R}^{d'}$ and define a self-supervised task \;
		\For{$t = 1,...,T$}
		{
		 Observe context features at time $t$: $x_{a,t} $ for each $a \in [N]$. \;
		 Get representation of context features at time $t$ using neural network $f(x_{a,t})$ for each $a \in [N]$. \;
		\For{all $a$ at time $t$}
		{
	   $\hat{r}_{a,t} = \hat{\theta}_t^Tf_t(x_{a,t}) $ \;
	   $ s_{a,t} = \sqrt{f_t(x_{a,t})^TD_a^{-1}f_t(x_{a,t})}$ \;
	   $p_{a,t} \leftarrow   \hat{r}_{a,t} + \alpha s_{a,t}$ \;
		}  \;
		Choose arm $ a_t = \arg \max p_{a,t}$, observe a real valued payoff $ r_{a_t,t}$ and update $ y_t$ .\;
		Update history  $H_t = (x_{a_t,t} , a_t , r_{a_t,t})$ \;
		\If{ $t \mod \beta == 0$}
		{
		   Retrain Neural Network $f_{t}$ using history $H_t$ and self-supervised task to get $f_{t+1}$    \;
		}
		\Else
		{
		$f_{t+1} = f_{t}$
		}
		Update $D_{a_t} = \sum_{\tau }  f_{t+1}(x_{a_\tau,\tau})  f_{t+1}(x_{a_\tau,\tau})^T $\;
		Update $b_{a_t} =\sum_{\tau }  r_{a_\tau,\tau} f_{t+1}(x_{a_\tau,\tau}) $\;
		Update $\hat{\theta}_{a_t} = D_{a_t}^{-1}b_{a_t}$\;
		\KwOut{Output: $ a_t $} \;
		}
\end{algorithm}
\begin{table}%[htbp!]
\begin{center}
\caption{$\ell_{\text{examples}}$ denotes the number of examples used for a contextual bandit problem, and  $N$ the number of actions  (classes in this case) in a given dataset;  and \#Rep  the number of repetitions with different initializations/seeds.}
\begin{tabular}{lrrr@{\qquad}ccc}
%         & $S_{train}$ & $S_{test}$ &$S_{classes}$  \\
& $\ell_{\text{train}}$ & $N$  & \textbf{\#Rep} &Image Size \\
\midrule
MNIST     &  60000     &  10  & 10 & $28 \times 28 \times 1$   \\
F-MNIST &  60000   &10 &  10    & $28 \times 28 \times 1$    \\
CIFAR-10 &   50000    & 10 &  10  & $32 \times32 \times 3$   \\   
CINIC-10 &   50000    & 10 &  10 & $32 \times32 \times 3$    \\ 
STL-10 &  5000   & 10 &  10 &$32 \times32 \times 3$  \\   
Intel &   14000    & 6 &  10  & $32 \times32 \times 3$ \\   
Food-10 &   10000    & 10 &  10    & $32 \times32 \times 3$ \\   
Imagenet-10 &   13000    & 10 &  10  &$32 \times32 \times 3$  \\   
\bottomrule
\end{tabular}
\label{tab:data_stats}
\end{center}
\end{table}

\section{Experiments}
Contextual bandit settings are very common in ad recommendations as the user visits the webpage. Image ads are a common ad type and image classification is essential for such ad recommendations.
We evaluate contextual bandits algorithm on image classification datasets. Converting multiclass classification datasets to bandits setup is a common way of evaluation in contextual bandits community because of lack of datasets with bandit feedback \cite{deshmukh2017multi,riquelme2018deep}. Each class in multiclass classification is considered as an arm or action. At every time step (iteration) $t$, agent is shown  a feature vector and the goal of agent is to recommend an arm (class) for that feature vector. If the agent predicts the arm (class label) correctly then it gets rewarded and if it predicts wrong it gets no reward. In this work, we focus on image classification because the primary aim is to get good feature representation in bandit tasks. For the self-supervision, the auxiliary task is rotation as defined in  \cite{gidaris2018unsupervised}.

\subsection{Datasets}
We use MNIST \cite{lecun1998gradient}, Fashion-MNIST (F-MNIST) \cite{xiao2017/online}, CIFAR-10 \cite{krizhevsky2009learning}, CINIC-10 \cite{darlow2018cinic}, STL-10 \cite{coates2011analysis}, Intel \cite{intel}, Food-10 \cite{bossard2014food}, and Imagenet-10 \cite{krizhevsky2012imagenet} datasets to demonstrate the effectiveness of the method proposed.  For MNIST and F-MNIST, we use all of the 60000 training examples. For CIFAR-10 and STL-10 we use all of the 50000 and 5000 training examples respectively. For CINIC-10, we randomly samply 50000 examples for each run from 90000 available training examples. For Food-10 and Imagenet-10, we select 10 classes out of 101 and 1000 classes in the original datasets respectively and then use all the training samples in those selected classes. For Intel dataset we use all of the 14000 training examples. Note that, in bandits setting there is no distinction made between training and test phase of the learning which is the case in traditional machine learning. In the table below \ref{tab:data_stats}, we provide the details of the datasets used. As noted in the table, we use ten random seeds and $N$ number of arms (classes). Using ten different random seeds leads to different sequence of images shown to the bandit solver in each of the ten different runs.

%$\epsilon$  the cooling parameter for self supervised loss,

MNIST contains images of numerical digits that are to be classified in their semantic number class ($ 0, \dots, 9 $). F-MNIST contains images of ten human dress types (classes) such as trouser, pullover, etc which are to be classified into their corresponding class labels. CIFAR-10, CINIC-10 and STL-10 contain images of animate (bird, car, deer, etc) and inanimate (airplane, truck, etc)  objects that are classified into their respective class labels given an image. Intel image classification dataset has images of natural scenes around the world such as buildings, forest, glaciers, etc. Images from these scenes are classified into their class label given an image from this data. Food-10 is a subset of original food-101 dataset \cite{bossard2014food}, where we randomly selected 10 classes out of the 101 classes for the classification task. The food-101 dataset contains images of 101 kinds of food that are to be classified to their corresponding class label given a food image. The selected classes are: Apple pie, Carrot cake, Chicken curry, Clam chowder, Falafel, Ice cream, Garlic bread, Mussels, Panckaes, Takoyaki. Imagenet-10 is a collection of ten randomly selected classes from the original Imagenet data with 1000 classes. The selected classes are: Stingray,  Thunder snake, Oystercatcher, Beaver, Baseball, Shower curtain, Toyshop, Trailer truck, Strawberry and Bubble. 

MNIST and F-MNIST are grey scale images. CIFAR-10, CINIC-10, STL-10, Intel, Food-10, and Imagenet-10 are all colored images. MNIST and F-MNIST are relatively larger datasets (60000 images) compared to CIFAR-10, CINIC-10, STL-10, Intel, Food-10, and Imagenet-10. The original MNIST and F-MNIST are $ 28 \times 28 \times 1 $ pixels whereas CIFAR-10, CINIC-10 are $ 32 \times 32 \times 3 $ pixels. STL-10 images contain $ 96 \times 96 \times 3 $ pixels,   Intel images contain $ 150 \times 150 \times 3 $ pixels,  Food-10 images contain $ 384 \times 384 \times 3$ pixels, and Imagenet images contain $ 256 \times 256 \times 3$ pixels per image. Apart from  MNIST and F-MNIST, images from other datasets were resized to $32 \times32 \times 3$. We observe that images from MNIST and F-MNIST have a simpler representation in terms of structure in the images compared to Imagenet or CIFAR-10 images. This inherent structure is exploited well by the self-supervision scheme proposed as we will observe in results section later. 

\subsection{Neural Network Implementation}
\begin{figure}
\centering
\includegraphics[height=6.5cm]{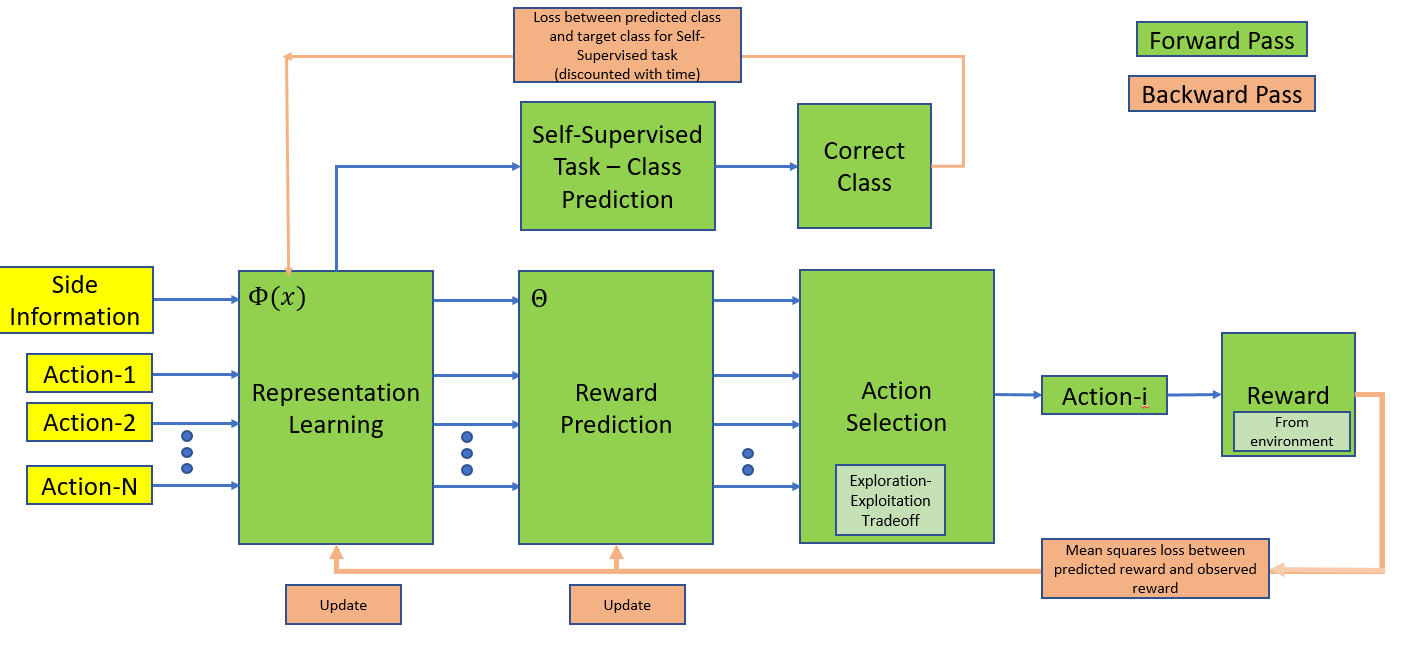}
\caption{Neural network block diagram for self-supervised contextual bandits}
\label{fig:SSneuralUCBArch}
\end{figure}

MNIST and F-MNIST use two layer convolutional neural network (CNN) and other datasets use Alexnet. Figure \ref{fig:SSneuralUCBArch} shows a block diagram of the implementation.  We use PyTorch for the neural network implementation. For MNIST and F-MNIST, we use stochastic gradient descent (SGD) with learning rate 0.25. For all other datasets we use Adam with learning rate 0.001. At the start of the experiment we pull each arm once to initialize the bandit problem as done traditionally \cite{riquelme2018deep,deshmukh2017multi}. We retrain the neural network for 10 epochs after every 500th arm-selection round (in Algorithm \ref{alg:SSNeural}, $ \beta = 500$). For all algorithms, we use $ \alpha = 1.0 $ and $ \gamma = 1.0 $.  MNIST, F-MNIST and CIFAR-10 use self-supervised loss weight $\mu = 0.9 $ and CINIC-10, Food-10, Intel, STL-10, and Imagenet-10 use self-supervised loss weight $\mu = 0.5$.

    \begin{figure}%[htbp]
    \centering
\sidesubfloat[]{\includegraphics[width=0.45\textwidth]{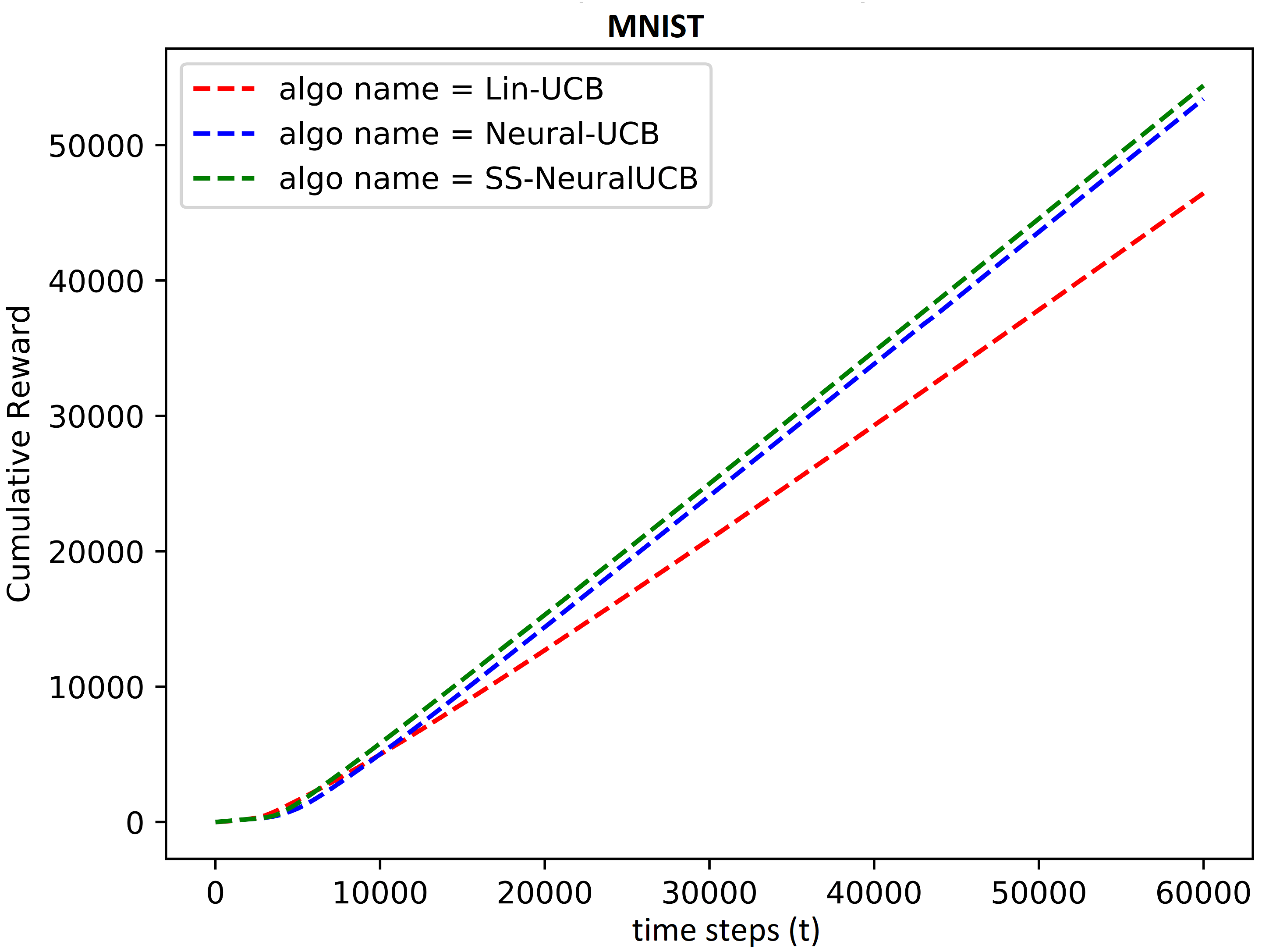}\label{fig:mnist}}
\hfil
\sidesubfloat[]{\includegraphics[width=0.45\textwidth]{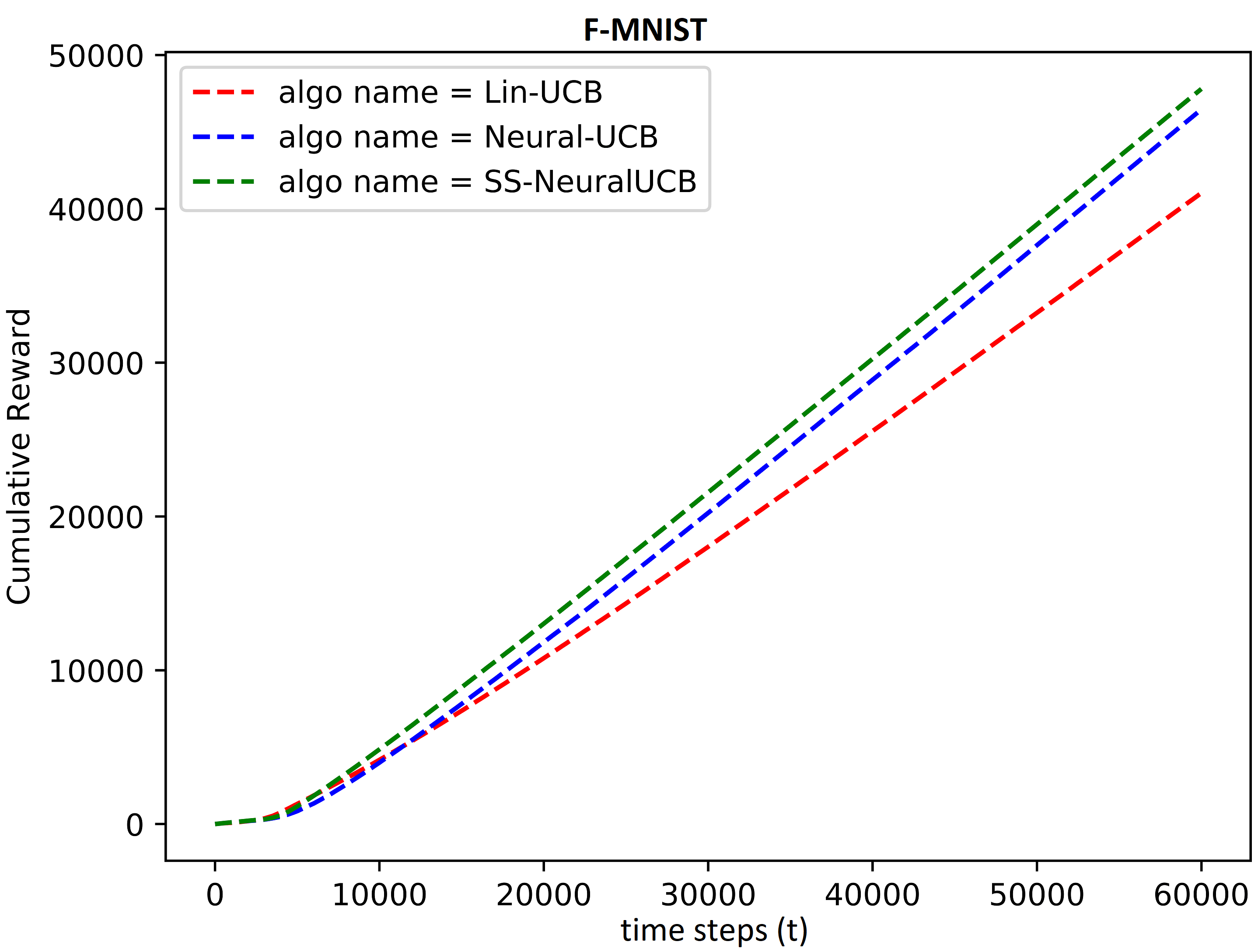}\label{fig:fmnist}}

\sidesubfloat[]{\includegraphics[width=0.45\textwidth]{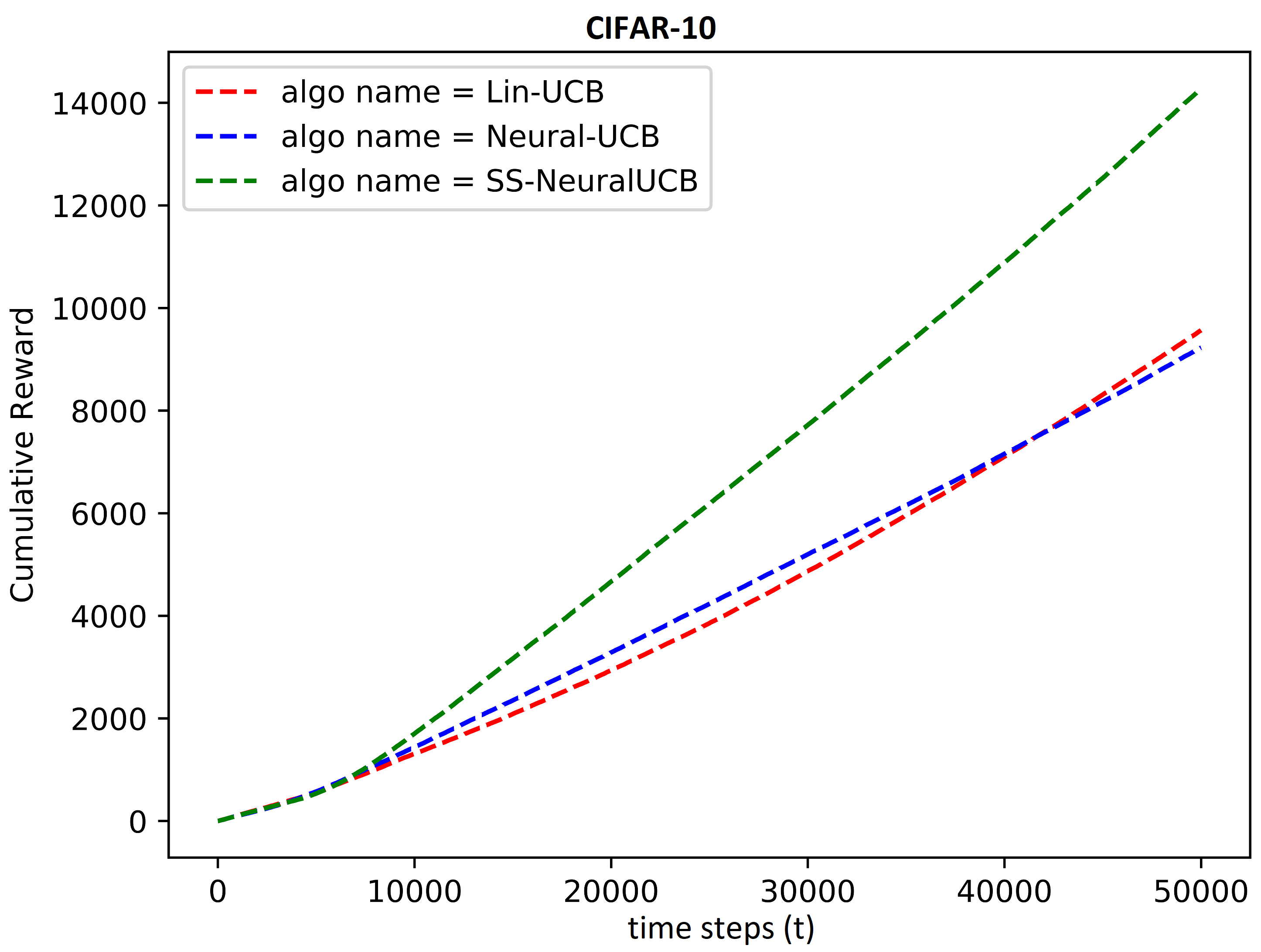}\label{fig:cifar10}}
\hfil
\sidesubfloat[]{\includegraphics[width=0.45\textwidth]{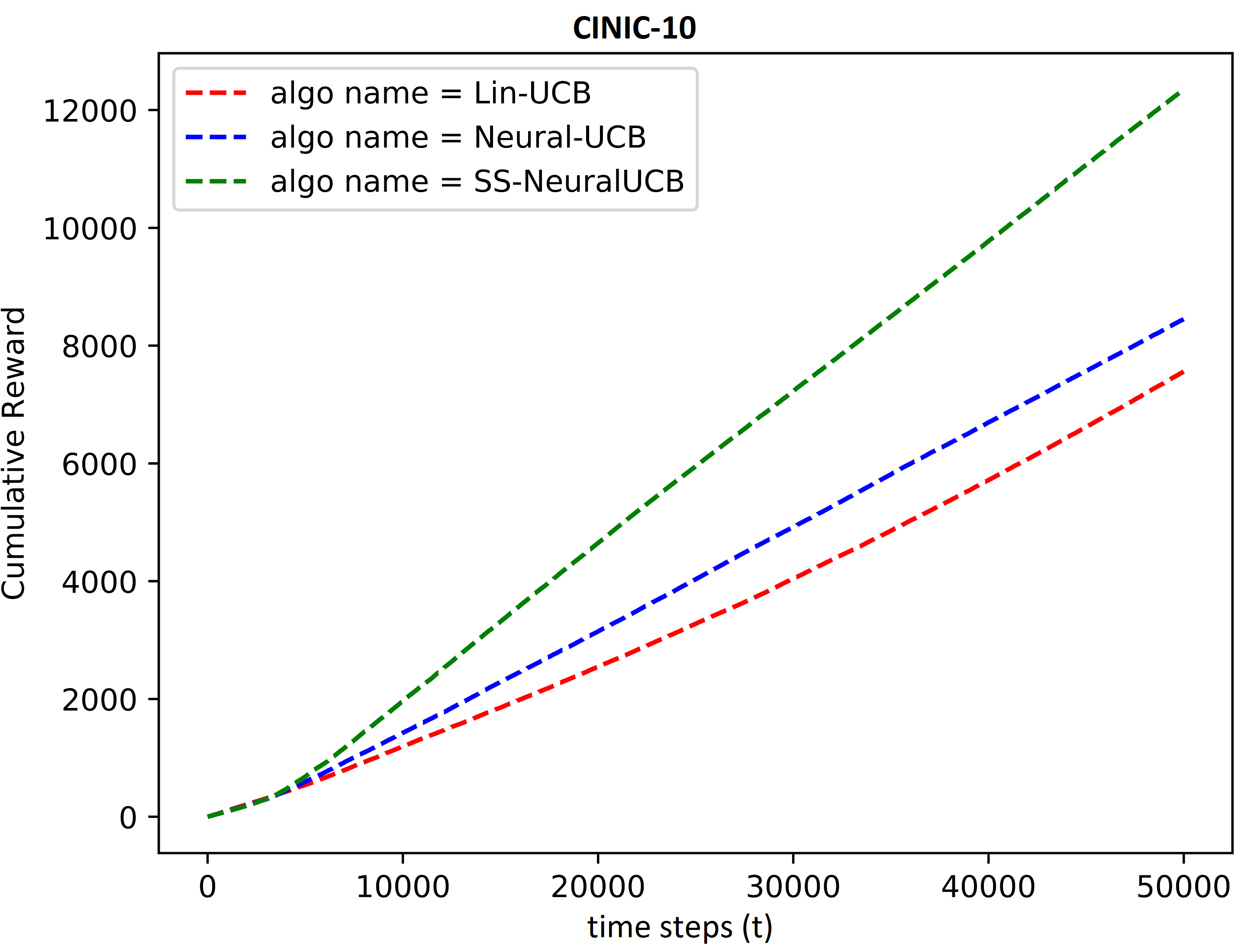}\label{fig:cinic10}}

\sidesubfloat[]{\includegraphics[width=0.45\textwidth]{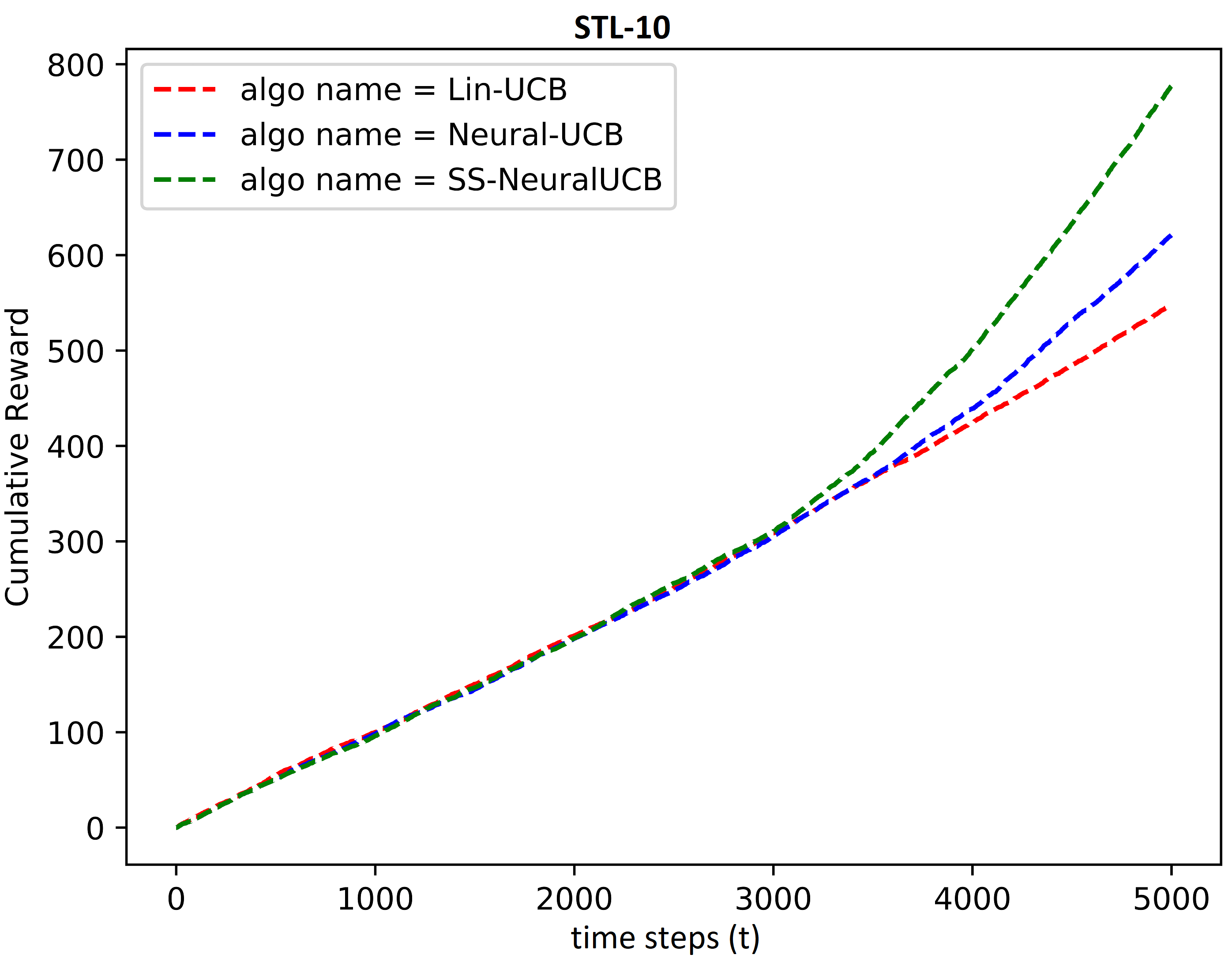}\label{fig:stl10}}
\hfil
\sidesubfloat[]{\includegraphics[width=0.45\textwidth]{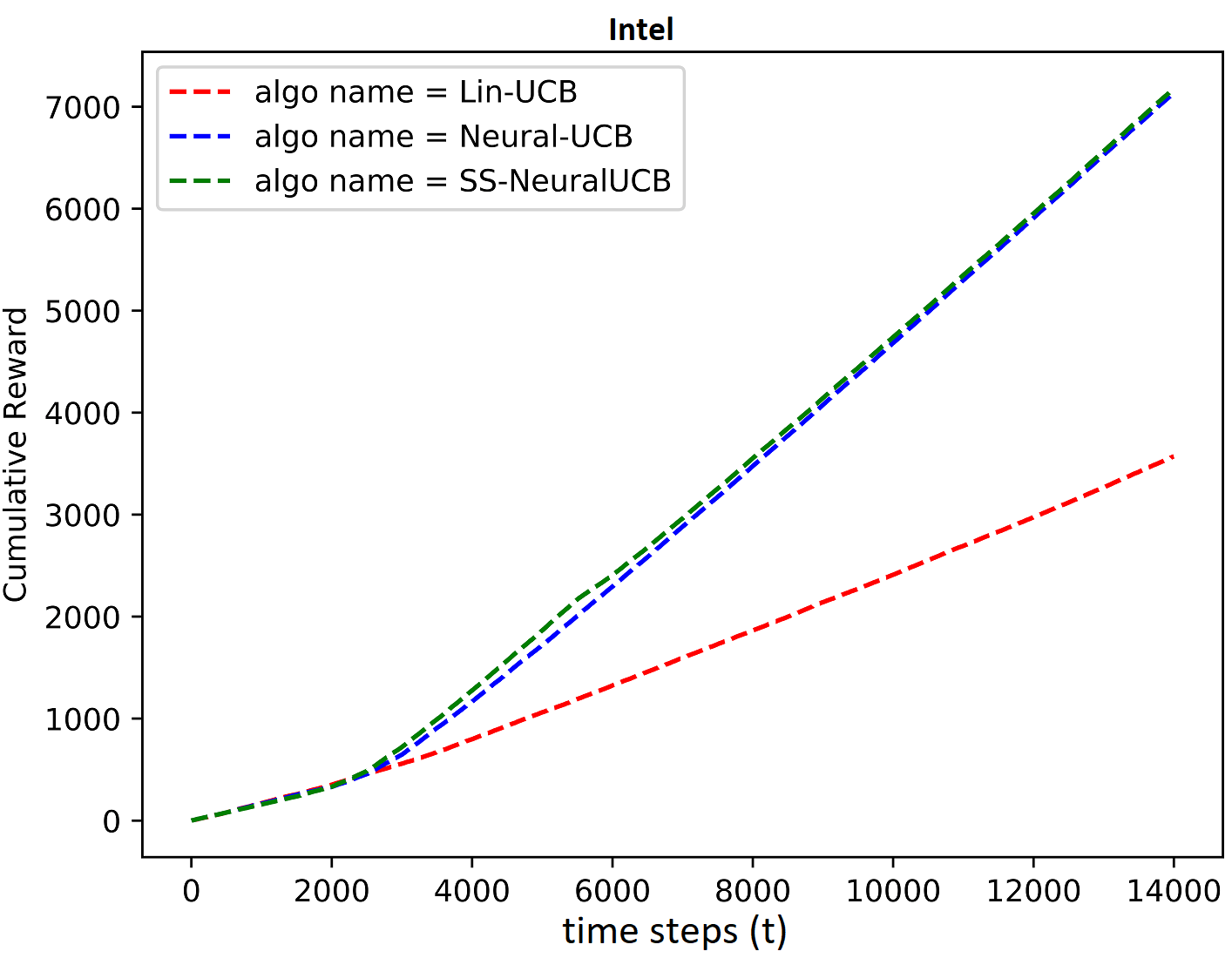}\label{fig:intel}}

\sidesubfloat[]{\includegraphics[width=0.45\textwidth]{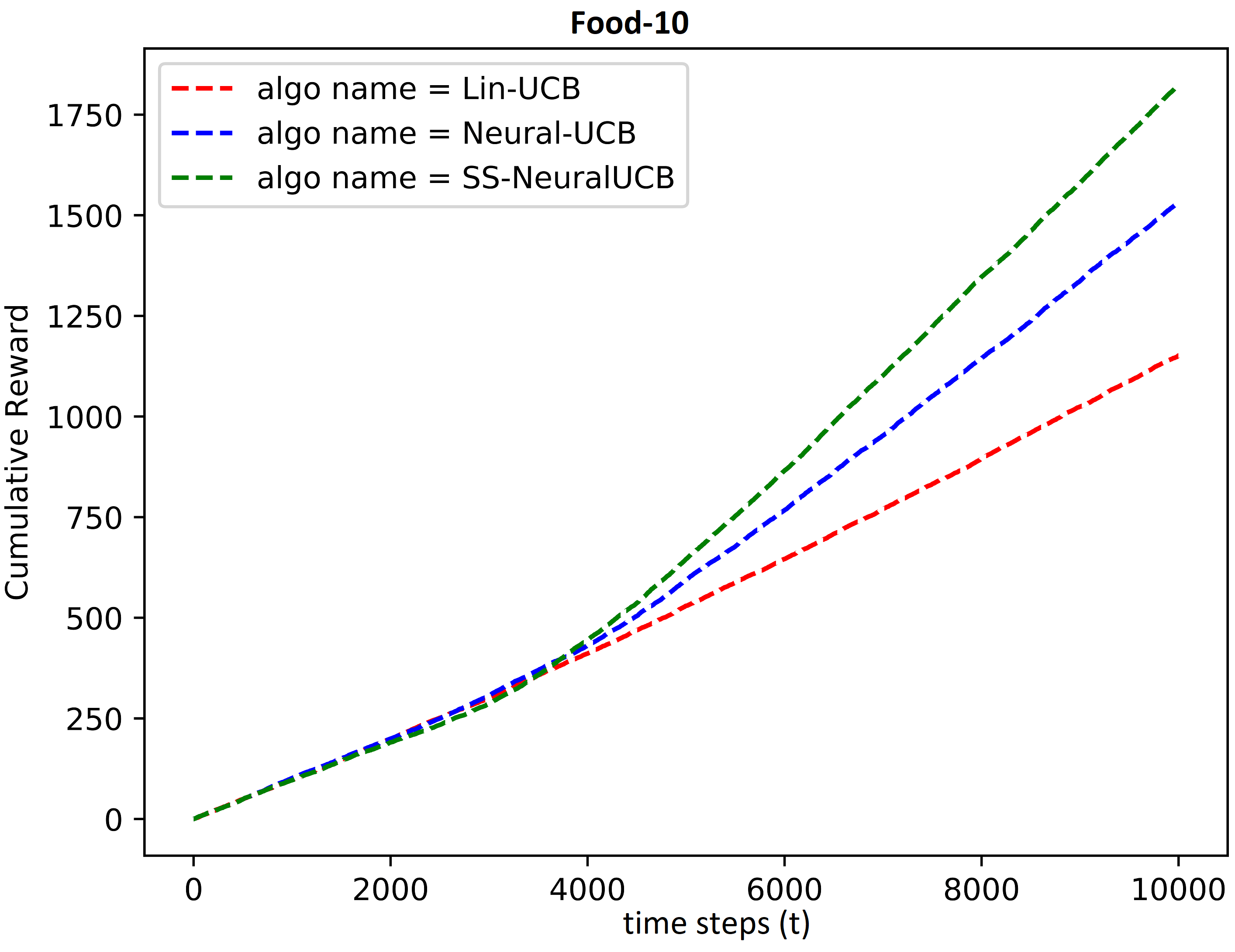}\label{fig:food}}
\hfil
\sidesubfloat[]{\includegraphics[width=0.45\textwidth]{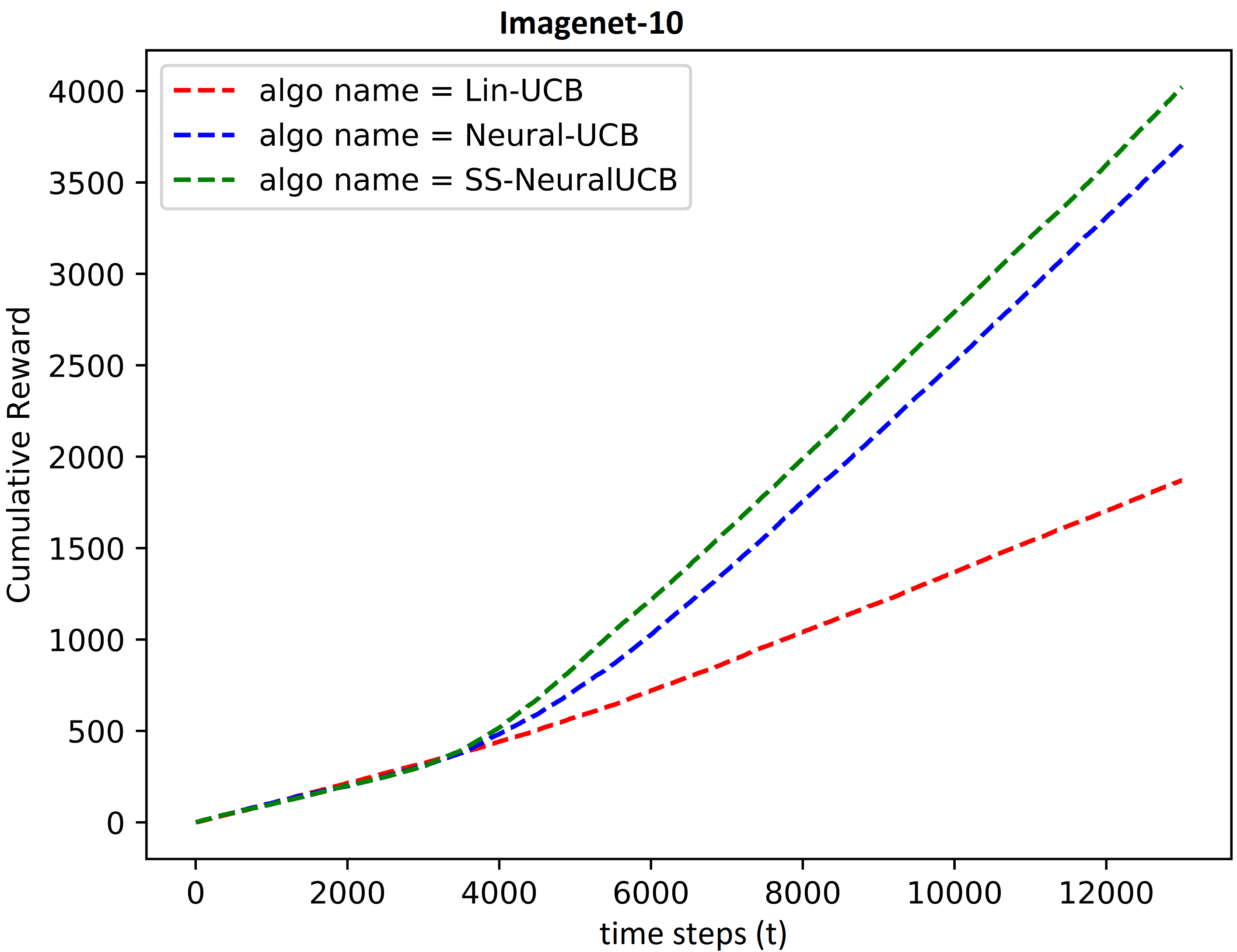}\label{fig:imagenet-10}}
\caption{Cumulative Reward plots}
 \label{fig:cumulative_reward}
 \end{figure}

\section{Results}
The overall results on all eight datasets are plotted and tabulated in Figure~\ref{fig:cumulative_reward} and Table~\ref{tab:cumulative_reward}, respectively. All the plots shown in Figure~\ref{fig:cumulative_reward} are averaged over ten random seed runs. The standard deviation of these results are reported in Table~\ref{tab:cumulative_reward} for each dataset and each approach. SS-Neural-UCB (Neural-UCB with self-supervision) outperforms all other methods across all eight datasets with wide margins in terms of rewards obtained. In the plots for MNIST and F-MNIST datasets (figure~\ref{fig:mnist}, and figure~\ref{fig:fmnist} respectively) we can clearly see that the SS-Neural-UCB peforms the best in terms of rewards over iterations (timesteps $t \in [1,\ldots,T]$ ) followed by Neural-UCB and Lin-UCB. The difference becomes more prominent with the increase in number of steps since the reward prediction becomes better as better feature representations are learnt over time.

 \begin{table}%[htbp]
\begin{center}
\caption{Mean cumulative reward for all eight datatsets over ten random seed runs---standard deviation reported in brackets. Higher mean and lower standard deviation is better.}
\begin{tabular}{lrrr}
& Lin$-$UCB & Neural$-$UCB  & SS$-$Neural$-$UCB\\
\midrule
MNIST     &  46444.8 ($\pm$ 141.33)    & 53404.2 ($\pm$219.55)  & 54395.2 ($\pm$ 129.50)    \\
F-MNIST & 41036.0 ($\pm$ 136.14)    & 46467.5 ($\pm$ 278.39)  & 47800.6 ($\pm$ 113.64)      \\
CIFAR-10 &    9567.4 ($\pm$ 108.35)    & 9227.8 ($\pm$ 1950.52)  & 14278.0 ($\pm$ 396.80)      \\   
CINIC-10 &    7560.1 ($\pm$ 72.88)    & 8449.7 ($\pm$ 951.46)  & 12346.7 ($\pm$ 418.38)    \\ 
STL-10 & 548.5 ($\pm$ 19.10)    & 621.1 ($\pm$ 52.47)  & 777.1 ($\pm$ 42.7)      \\   
Intel &  3571.5 ($\pm$ 124.51)    &7136.2 ($\pm$ 226.90)  & 7178.8 ($\pm$ 1285.44)  \\   
Food-10 &  1151.4 ($\pm$40.71)    & 1533.0 ($\pm$ 91.1)  & 1823.1 ($\pm$ 56.13)    \\   
Imagenet-10 &  1870.8 ($\pm$ 45.21)    & 3705.5 ($\pm$ 259.11)  & 4021.4 ($\pm$ 726.32)     \\   
\bottomrule
\end{tabular}
\label{tab:cumulative_reward}
\end{center}
\end{table}

\begin{table}%[htbp]
\begin{center}
\caption{Number of times each of the algorithm was ranked top in terms of cumulative reward}
\begin{tabular}{lrrr@{\qquad}cc}
& Lin-UCB & Neural-UCB  & SS-Neural-UCB\\
\midrule
MNIST     &  0  & 0 & 10   \\
F-MNIST & 0 & 0 & 10      \\
CIFAR-10 &  0 &  0 & 10        \\   
CINIC-10 &  0     & 0  & 10        \\ 
STL-10 &  0  & 0 & 10      \\   
Intel &   0 & 2 & 8     \\   
Food-10 &  0  & 0 & 10  \\   
Imagenet-10 &0  & 1 & 9      \\   
\bottomrule
\end{tabular}
\label{tab:ranking}
\end{center}
\end{table}

\begin{table}%[htbp]
\begin{center}
\caption{Absolute percentage gain in cumulative rewards by SS-Neural-UCB over the next best bandits solver (Neural-UCB).}
\begin{tabular}{lrrr@{\qquad}cc}
Dataset & Gain by SS-Neural-UCB (in \%)\\
\midrule
MNIST     &   1.86   \\
F-MNIST &   2.87     \\
CIFAR-10 &  54.7        \\   
CINIC-10 &  46.11       \\ 
STL-10 &  24.1       \\   
Intel &   0.50    \\   
Food-10 &  18.92 \\   
Imagenet-10 &  8.53  \\   
\bottomrule
\end{tabular}
\label{tab:percent_gain}
\end{center}
\end{table}

\subsection{Benefits of Self-Supervision }
Figures~\ref{fig:stl10},~\ref{fig:imagenet-10},~\ref{fig:cifar10}, and~\ref{fig:food} show the reward vs timesteps plots for STL-10, Imagenet-10, CIFAR-10, and Food-10 datasets. Similar to MNIST and F-MNIST results we see the SS-Neural-UCB outperforms all other methods. Additionally, the performance gap between best (SS-Neural-UCB) and the second best method (Neural-UCB) gets wider as timesteps increase. This can be explained by our earlier assertion that STL-10, Imagenet-10, CINIC-10, CIFAR-10, and Food-10 images have more complex representations. The self-supervision, by extracting these rich feature representations, helps the bandit solver achieve better reward prediction. STL-10, Food-10, and Imagenet-10 plots (figures~\ref{fig:stl10},~\ref{fig:food}, and~\ref{fig:imagenet-10}) show that even with fewer timesteps, i.e. fewer datapoints in bandit learning (5000 or less),  the SS-Neural-UCB is able to outperform all the other methods. This shows that self-supervision is learning good feature representation despite the usual lack of datapoints in the early phases of bandit solvers. This is an important attribute of this learning scheme as this helps to reach the optimal policy quickly.

\begin{figure}%[htbp]
\centering

\sidesubfloat[]{\includegraphics[width=0.4\textwidth]{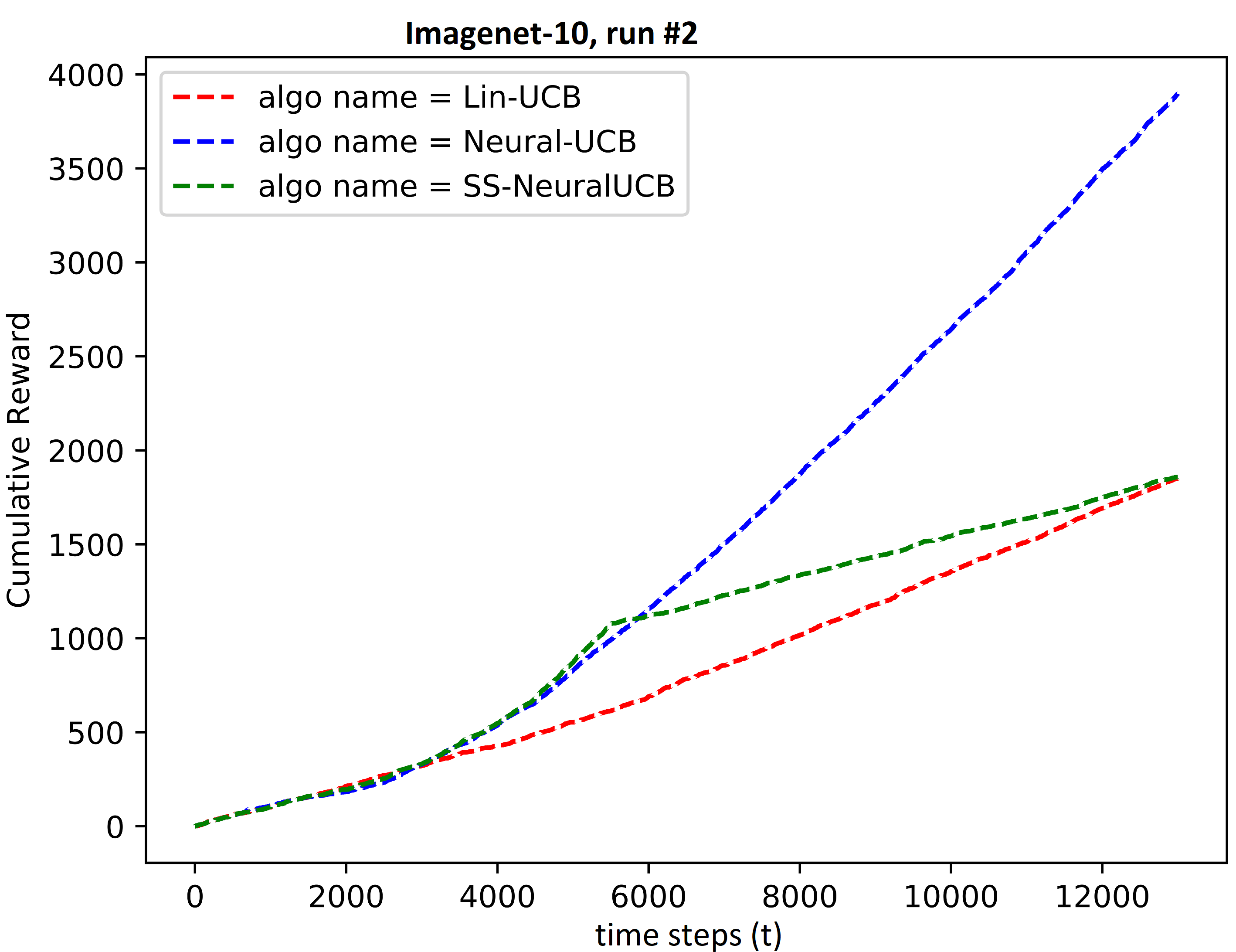}\label{fig:failure_imagenet}}

\medskip
\sidesubfloat[]{\includegraphics[width=0.4\textwidth]{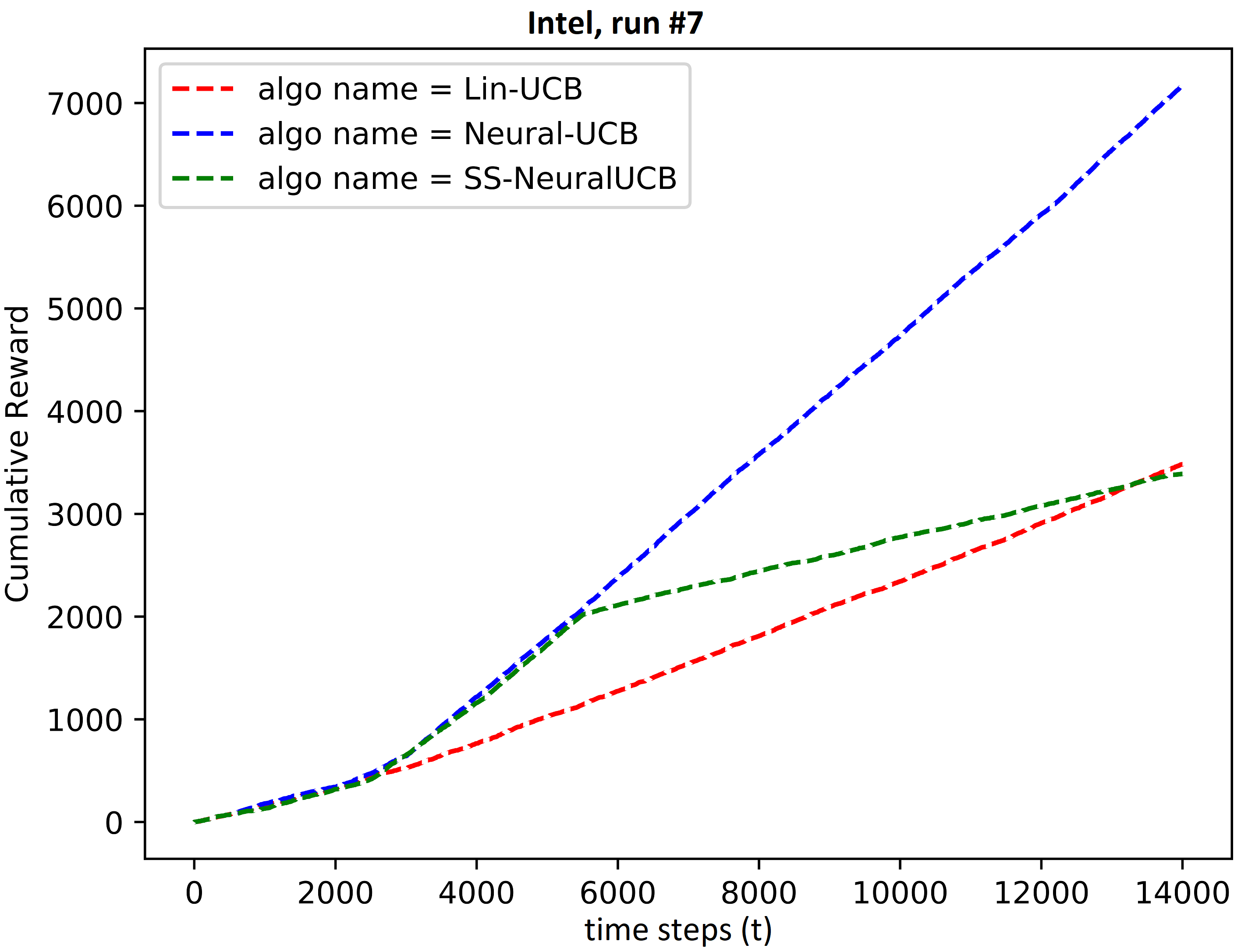}\label{fig:failure_intel6run7}}
\hfil
\sidesubfloat[]{\includegraphics[width=0.4\textwidth]{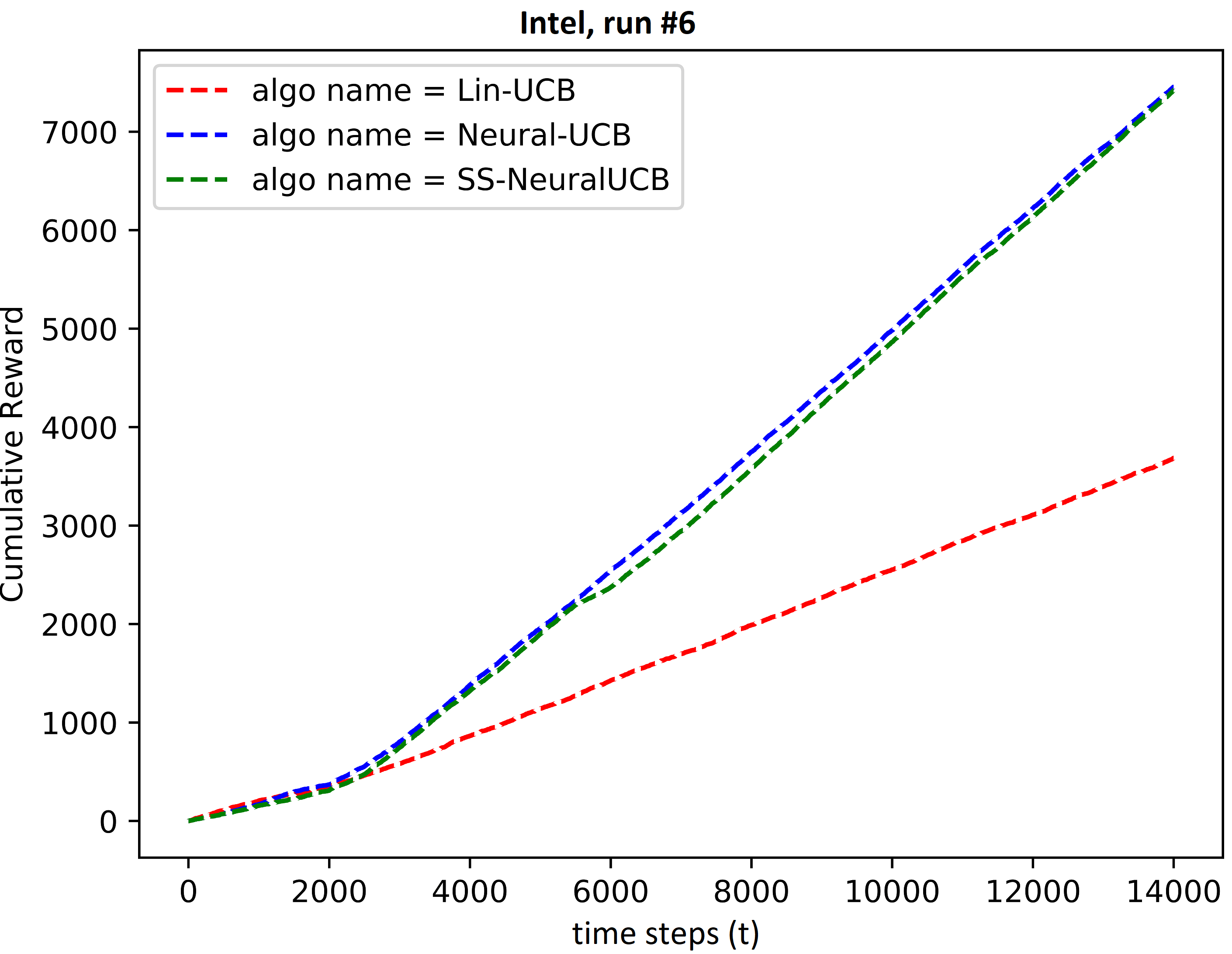}\label{fig:failure_intel6run6}}

\caption{Cumulative Reward plots for three out of eighty cases in the main results where proposed method SS-Neural-UCB is not the best method.}
 \label{fig:failure}
 \end{figure}

\subsection{Overall Performance}
Table~\ref{tab:ranking} tabulates statistics of the overall performance of the three bandit solvers over all eight datasets. It provides the number of times SS-Neural-UCB outperforms its competition across the eight different datasets and the ten random seed runs . We see that for all except two datasets, Intel and Imagenet, SS-Neural-UCB bests its closest second in all ten runs. For Intel dataset it performs better than the second best in eight trials whereas for Imagenet-10 it performs better in nine trials out of ten. Table~\ref{tab:cumulative_reward} shows the mean cumulative rewards obtained by the methods on the eight datasets along with the standard deviation over the ten runs. We see that we are better in the cumulative reward means as well as standard deviations across all eight datasets except two---Intel and Imagenet-10. For Intel and Imagenet-10 we outperform in the cumulative reward  means but not in the standard deviations. This case is also captured by the Table~\ref{tab:ranking} wherein we fail to outperform for the three runs in the two datasets. Table \ref{tab:percent_gain} shows the absolute percentage gain by the top performing method, i.e. the proposed method SS-Neural-UCB, over it's next best competitor (Neural-UCB). We can see that the gains obtained in datasets with rich structure (STL-10, Imagenet-10, CINIC-10, CIFAR-10, and Food-10) is quite high. The best performance comes on CIFAR-10 dataset (54\% gain) closely followed by CINIC-10 dataset (46\% gain).

\subsection{Suboptimal Self-Supervision}
As discussed earlier out of total eighty random seed runs (ten runs for each of the eight datasets), we fail to outperform the competing methods in three of them. Of these three cases, two belong to Intel dataset and one belongs to Imagenet-10, as shown in figure~\ref{fig:failure}. SS-Neural-UCB starts degrading rapidly over timesteps for figures~\ref{fig:failure_intel6run7} and~\ref{fig:failure_imagenet}. We suspect that few of the random seed runs lead to sequence of images in the beginning which is not a good sample of representation of overall dataset. This in turn leads to the self-supervision learning a bad representation in the beginning of the learning phase. Overall, this failure probability is very low: three out of eighty random runs i.e. less than 4\%. This is remarkably good given the variety of image datasets the bandit solver has to work with---from clothes (F-MNIST) to sceneries (Intel) and dishes (Food-10) to wild animals (Imagenet-10). We provide more suboptimal cases by giving higher weights to the self-supervision loss in the appendix.

\section{Conclusion}

We observe significant gains in cumulative reward for contextual bandits by leveraging the rich representation extracted via self-supervision. Out of 80 runs across 8 tasks the combined proposed approach outperforms in 77 of them---a success rate above 96\%. This is very promising given the diverse nature of image datasets. The insights gained from the results can lead to designing of bandit solvers that leverage true feature representation of the data leading to high gains in cumulative rewards. This is because the three failure cases are mostly due to the fact that we have a non representative sample sequence which forces self-supervision to learn sub-optimal feature representation.

The self-supervision starts helping from the get-go as seen in STL-10, Food-10, and Imagenet-10 plots (figures~\ref{fig:stl10},~\ref{fig:food}, and~\ref{fig:imagenet-10}). This encourages practitioners to use self-supervision in bandits more since bandits are a slow learner due to the fact that feedback (data) is much more scarce in the beginning phases of learning. Which self-supervised tasks help under what bandit settings is a fruitful area of research. This can lead to further optimized solvers that can find optimal policy relative quickly.

\section{Appendix}
\subsection{Too much Self-Supervision hurts}
\begin{figure}[htbp]
    \centering
\sidesubfloat[]{\includegraphics[width=0.45\textwidth]{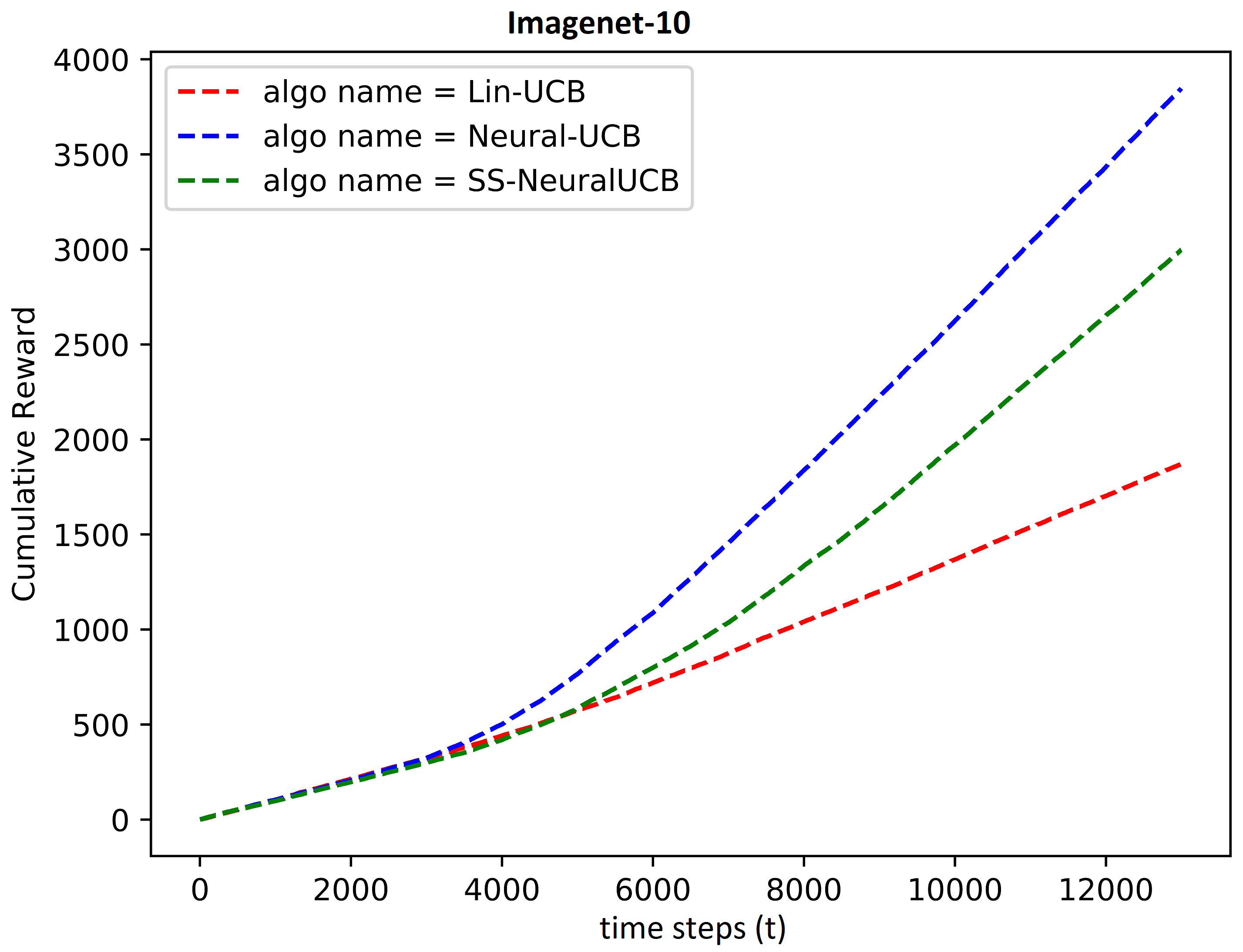}\label{fig:neg_imagenet_res}}
\hfil
\sidesubfloat[]{\includegraphics[width=0.45\textwidth]{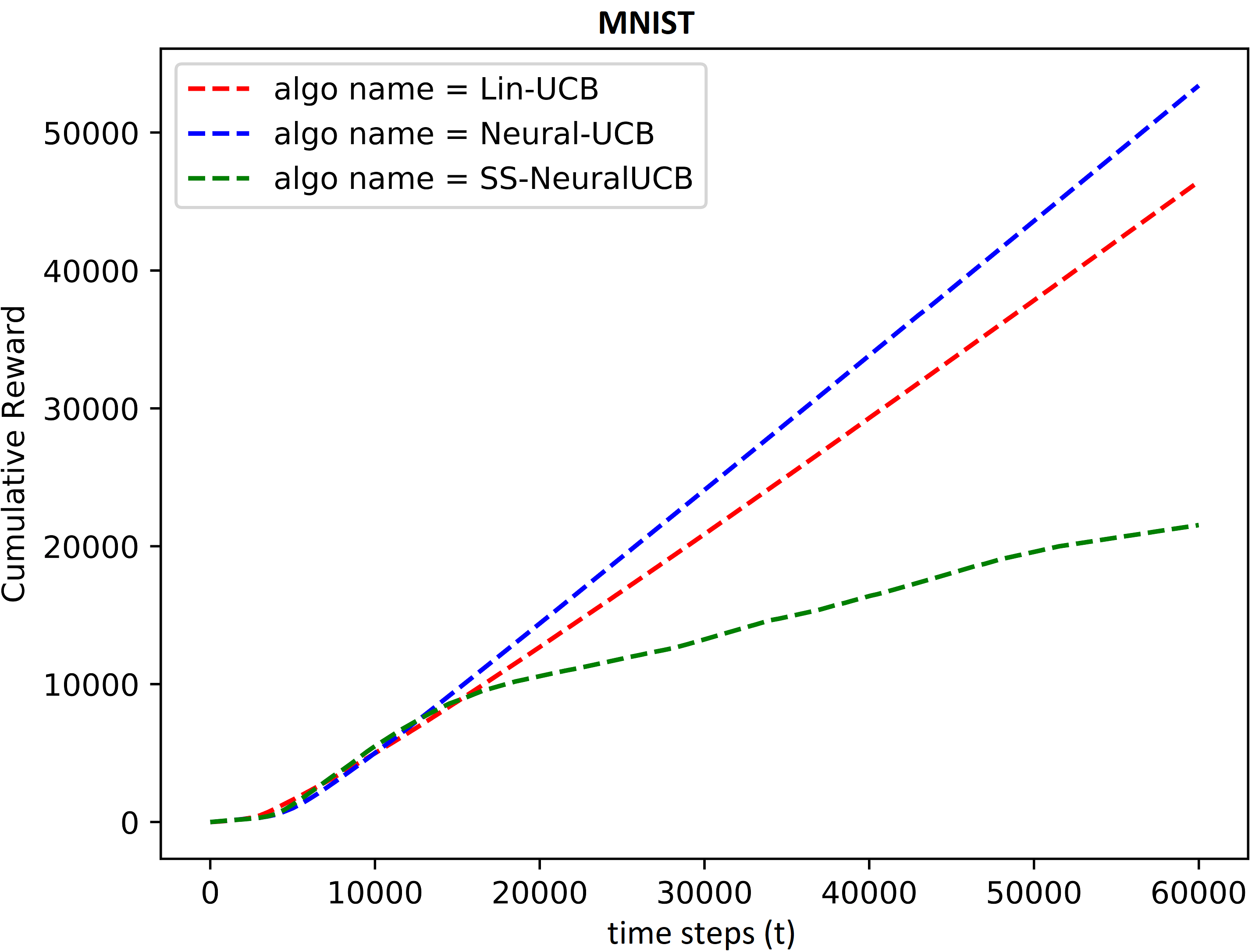}\label{fig:neg_mnist_res}}

\caption{Cumulative Reward plots for failure cases when self supervised loss is not decreased with time}
 \label{fig:neg_res}
 \end{figure}

Note that if one does not reduce the weight of self-supervision loss with time (i.e. $\mu = 1$ ), one gets very poor performance. The setting of  $\mu = 1$ is like a multi-task learning setting which is not really helpful in the case of contextual bandits. This fact is evident from the figures shown in \ref{fig:neg_res} 

\clearpage
\bibliographystyle{splncs04}
\bibliography{uai}

\begin{thebibliography}{10}
\providecommand{\url}[1]{\texttt{#1}}
\providecommand{\urlprefix}{URL }
\providecommand{\doi}[1]{https://doi.org/#1}

\bibitem{agrawal2013thompson}
Agrawal, S., Goyal, N.: Thompson sampling for contextual bandits with linear
  payoffs. In: International Conference on Machine Learning. pp. 127--135
  (2013)

\bibitem{bossard2014food}
Bossard, L., Guillaumin, M., Van~Gool, L.: Food-101--mining discriminative
  components with random forests. In: European conference on computer vision.
  pp. 446--461. Springer (2014)

\bibitem{burda2018large}
Burda, Y., Edwards, H., Pathak, D., Storkey, A., Darrell, T., Efros, A.A.:
  Large-scale study of curiosity-driven learning. arXiv preprint
  arXiv:1808.04355  (2018)

\bibitem{Caron_2018_ECCV}
Caron, M., Bojanowski, P., Joulin, A., Douze, M.: Deep clustering for
  unsupervised learning of visual features. In: The European Conference on
  Computer Vision (ECCV) (September 2018)

\bibitem{chu2011contextual}
Chu, W., Li, L., Reyzin, L., Schapire, R.: Contextual bandits with linear
  payoff functions. In: Proceedings of the Fourteenth International Conference
  on Artificial Intelligence and Statistics. pp. 208--214 (2011)

\bibitem{coates2011analysis}
Coates, A., Ng, A., Lee, H.: An analysis of single-layer networks in
  unsupervised feature learning. In: Proceedings of the fourteenth
  international conference on artificial intelligence and statistics. pp.
  215--223 (2011)

\bibitem{darlow2018cinic}
Darlow, L.N., Crowley, E.J., Antoniou, A., Storkey, A.J.: Cinic-10 is not
  imagenet or cifar-10. arXiv preprint arXiv:1810.03505  (2018)

\bibitem{deshmukh2017multi}
Deshmukh, A.A., Dogan, U., Scott, C.: Multi-task learning for contextual
  bandits. In: Advances in neural information processing systems. pp.
  4848--4856 (2017)

\bibitem{deshmukh2018simple}
Deshmukh, A.A., Sharma, S., Cutler, J.W., Moldwin, M., Scott, C.: Simple regret
  minimization for contextual bandits. arXiv preprint arXiv:1810.07371  (2018)

\bibitem{dosovitskiy2015discriminative}
Dosovitskiy, A., Fischer, P., Springenberg, J.T., Riedmiller, M., Brox, T.:
  Discriminative unsupervised feature learning with exemplar convolutional
  neural networks. IEEE transactions on pattern analysis and machine
  intelligence  \textbf{38}(9),  1734--1747 (2015)

\bibitem{Gidaris_2019_ICCV}
Gidaris, S., Bursuc, A., Komodakis, N., Perez, P., Cord, M.: Boosting few-shot
  visual learning with self-supervision. In: The IEEE International Conference
  on Computer Vision (ICCV) (October 2019)

\bibitem{gidaris2018unsupervised}
Gidaris, S., Singh, P., Komodakis, N.: Unsupervised representation learning by
  predicting image rotations. arXiv preprint arXiv:1803.07728  (2018)

\bibitem{goyal2019scaling}
Goyal, P., Mahajan, D., Gupta, A., Misra, I.: Scaling and benchmarking
  self-supervised visual representation learning. In: Proceedings of the IEEE
  International Conference on Computer Vision. pp. 6391--6400 (2019)

\bibitem{intel}
Intel: Intel scene classification challenge.
  https://datahack.analyticsvidhya.com/contest/practice-problem-intel-scene-classification-challe/
   (2018)

\bibitem{pmlr-v22-kaufmann12}
Kaufmann, E., Cappe, O., Garivier, A.: On bayesian upper confidence bounds for
  bandit problems. In: Lawrence, N.D., Girolami, M. (eds.) Proceedings of the
  Fifteenth International Conference on Artificial Intelligence and Statistics.
  Proceedings of Machine Learning Research, vol.~22, pp. 592--600. PMLR, La
  Palma, Canary Islands (21--23 Apr 2012)

\bibitem{krizhevsky2012imagenet}
Krizhevsky, A., Sutskever, I., Hinton, G.E.: Imagenet classification with deep
  convolutional neural networks. In: Advances in neural information processing
  systems. pp. 1097--1105 (2012)

\bibitem{krizhevsky2009learning}
Krizhevsky, A., et~al.: Learning multiple layers of features from tiny images
  (2009)

\bibitem{kveton2019garbage}
Kveton, B., Szepesvari, C., Vaswani, S., Wen, Z., Lattimore, T., Ghavamzadeh,
  M.: Garbage in, reward out: Bootstrapping exploration in multi-armed bandits.
  In: International Conference on Machine Learning. pp. 3601--3610 (2019)

\bibitem{lecun1998gradient}
LeCun, Y., Bottou, L., Bengio, Y., Haffner, P.: Gradient-based learning applied
  to document recognition. Proceedings of the IEEE  \textbf{86}(11),
  2278--2324 (1998)

\bibitem{li2011unbiased}
Li, L., Chu, W., Langford, J., Wang, X.: Unbiased offline evaluation of
  contextual-bandit-based news article recommendation algorithms. In:
  Proceedings of the fourth ACM international conference on Web search and data
  mining. pp. 297--306 (2011)

\bibitem{li2016collaborative}
Li, S., Karatzoglou, A., Gentile, C.: Collaborative filtering bandits. In:
  Proceedings of the 39th International ACM SIGIR conference on Research and
  Development in Information Retrieval. pp. 539--548 (2016)

\bibitem{pami/LiuWB19}
Liu, X., van~de Weijer, J., Bagdanov, A.D.: Exploiting unlabeled data in cnns
  by self-supervised learning to rank. {IEEE} Trans. Pattern Anal. Mach.
  Intell.  \textbf{41}(8),  1862--1878 (2019).
  \doi{10.1109/TPAMI.2019.2899857},
  \url{https://doi.org/10.1109/TPAMI.2019.2899857}

\bibitem{McMahan_35401}
McMahan, H.B., Streeter, M.: Tighter bounds for multi-armed bandits with expert
  advice. In: Proceedings of the 22nd Annual Conference on Learning Theory
  {(COLT)} (2009)

\bibitem{mikolov2013efficient}
Mikolov, T., Chen, K., Corrado, G., Dean, J.: Efficient estimation of word
  representations in vector space. arXiv preprint arXiv:1301.3781  (2013)

\bibitem{noroozi2016unsupervised}
Noroozi, M., Favaro, P.: Unsupervised learning of visual representations by
  solving jigsaw puzzles. In: European Conference on Computer Vision. pp.
  69--84. Springer (2016)

\bibitem{pathak2017curiosity}
Pathak, D., Agrawal, P., Efros, A.A., Darrell, T.: Curiosity-driven exploration
  by self-supervised prediction. In: Proceedings of the IEEE Conference on
  Computer Vision and Pattern Recognition Workshops. pp. 16--17 (2017)

\bibitem{pathak2019self}
Pathak, D., Gandhi, D., Gupta, A.: Self-supervised exploration via
  disagreement. arXiv preprint arXiv:1906.04161  (2019)

\bibitem{10.1145/1645953.1646072}
Qiu, L., Zhang, W., Hu, C., Zhao, K.: Selc: A self-supervised model for
  sentiment classification. In: Proceedings of the 18th ACM Conference on
  Information and Knowledge Management. p. 929–936. CIKM ’09, Association
  for Computing Machinery, New York, NY, USA (2009).
  \doi{10.1145/1645953.1646072}, \url{https://doi.org/10.1145/1645953.1646072}

\bibitem{rebuffi2019semisupervised}
Rebuffi, S.A., Ehrhardt, S., Han, K., Vedaldi, A., Zisserman, A.:
  Semi-supervised learning with scarce annotations (2019)

\bibitem{riquelme2018deep}
Riquelme, C., Tucker, G., Snoek, J.: Deep bayesian bandits showdown: An
  empirical comparison of bayesian deep networks for thompson sampling. arXiv
  preprint arXiv:1802.09127  (2018)

\bibitem{saunshi2019theoretical}
Saunshi, N., Plevrakis, O., Arora, S., Khodak, M., Khandeparkar, H.: A
  theoretical analysis of contrastive unsupervised representation learning. In:
  International Conference on Machine Learning. pp. 5628--5637 (2019)

\bibitem{Singh2018SelfSupervisedFL}
Singh, S., Batra, A., Pang, G., Torresani, L., Basu, S., Paluri, M., Jawahar,
  C.V.: Self-supervised feature learning for semantic segmentation of overhead
  imagery. In: BMVC (2018)

\bibitem{srinivas2010gaussian}
Srinivas, N., Krause, A., Kakade, S., Seeger, M.: Gaussian process optimization
  in the bandit setting: no regret and experimental design. In: Proceedings of
  the 27th International Conference on International Conference on Machine
  Learning. pp. 1015--1022 (2010)

\bibitem{taha2018stream}
Taha, A., Meshry, M., Yang, X., Chen, Y.T., Davis, L.: Two stream
  self-supervised learning for action recognition (2018)

\bibitem{tewari2017ads}
Tewari, A., Murphy, S.A.: From ads to interventions: Contextual bandits in
  mobile health. In: Mobile Health, pp. 495--517. Springer (2017)

\bibitem{Tokic_10_1007}
Tokic, M., Palm, G.: Value-difference based exploration: Adaptive control
  between epsilon-greedy and softmax. In: Bach, J., Edelkamp, S. (eds.) KI
  2011: Advances in Artificial Intelligence. pp. 335--346. Springer Berlin
  Heidelberg, Berlin, Heidelberg (2011)

\bibitem{valko2013finite}
Valko, M., Korda, N., Munos, R., Flaounas, I., Cristianini, N.: Finite-time
  analysis of kernelised contextual bandits. In: Uncertainty in Artificial
  Intelligence. p.~654. Citeseer (2013)

\bibitem{xiao2017/online}
Xiao, H., Rasul, K., Vollgraf, R.: Fashion-mnist: a novel image dataset for
  benchmarking machine learning algorithms (2017)

\bibitem{zhang2016colorful}
Zhang, R., Isola, P., Efros, A.A.: Colorful image colorization. In: European
  conference on computer vision. pp. 649--666. Springer (2016)

\end{thebibliography}
\end{document}